\pdfoutput=1

\documentclass[11pt]{article}

\usepackage[]{acl} 
\usepackage{times}
\usepackage{latexsym}
\usepackage{booktabs}
\usepackage{color, colortbl, xcolor}
\usepackage{inconsolata}
\usepackage{subcaption}
\usepackage{enumitem}
\usepackage[most]{tcolorbox}
\usepackage[T1]{fontenc}
\usepackage[main=english,russian]{babel}
\usepackage[utf8]{inputenc}
\usepackage{microtype}
\usepackage{inconsolata}
\usepackage{graphicx}
\usepackage{hyperref}
\usepackage{cleveref}
\usepackage{placeins}
\usepackage{float}
\usepackage{multirow}
\usepackage{longtable}

\definecolor{Lightgreen}{RGB}{144,238,144}
\definecolor{Gray}{gray}{0.9}
\definecolor{grey}{gray}{0.9}

\crefformat{section}{\S#2#1#3} 
\crefformat{subsection}{\S#2#1#3}
\crefformat{subsubsection}{\S#2#1#3}

\newcommand{\dataset}{\textsf{\small SynthDetoxM}}

\tcbset{
  colback=gray!10,        
  colframe=black!30,      
  width=\linewidth,       
  boxrule=0.4mm,          
  arc=2mm,                
  outer arc=2mm,          
  boxsep=5mm,             
}

\title{SynthDetoxM: Modern LLMs are Few-Shot \\ Parallel Detoxification Data Annotators}

\author{
  Daniil Moskovskiy\textsuperscript{1,2}\thanks{Equal contribution.} \quad
  Nikita Sushko\textsuperscript{1,2}\footnotemark[1] \quad
  Sergey Pletenev\textsuperscript{1,2} \quad \\
  \textbf{Elena Tutubalina\textsuperscript{1,3,4}} \quad
  \textbf{Alexander Panchenko\textsuperscript{2,1}} \\ 
  \textsuperscript{1}AIRI~~~
  \textsuperscript{2}Skoltech~~~ 
  \textsuperscript{3}Sber AI~~~ 
  \textsuperscript{4}ISP RAS Research Center for Trusted AI \\
  \\
  \small{\textbf{Correspondence:} \href{mailto:D.Moskovskiy@skol.tech}{\{d.moskovskiy, a.panchenko\}@skol.tech}}
}

\begin{document}
\maketitle
\begin{abstract}
Existing approaches to multilingual text detoxification are hampered by the scarcity of parallel multilingual datasets. In this work, we introduce a pipeline for the generation of multilingual parallel detoxification data. We also introduce \dataset, a manually collected and synthetically generated multilingual parallel text detoxification dataset comprising 16,000 high-quality detoxification sentence pairs across German, French, Spanish and Russian. The data was sourced from different toxicity evaluation datasets and then rewritten with nine modern open-source LLMs in few-shot setting. Our experiments demonstrate that models trained on the produced synthetic datasets have superior performance to those trained on the human-annotated MultiParaDetox dataset even in data limited setting. Models trained on \dataset~outperform all evaluated LLMs in few-shot setting. We release our dataset and code to help further research in multilingual text detoxification.

\textcolor{red}{\footnotesize Warning: this paper contains illustrative examples of texts that readers may find offensive or disturbing.}
\end{abstract}

\section{Introduction}

The proliferation of social networks and text-based internet media has highlighted the issue of online toxicity and hate speech~\cite{Saha2019PrevalenceAP}. This phenomenon not only creates an unpleasant environment for users but also deters advertisers, potentially impacting the economic viability of these platforms~\cite{fortuna2018survey}. Consequently, there is an urgent need for effective mechanisms to measure and mitigate toxicity in online spaces.

A promising approach to addressing this challenge is text detoxification of text through paraphrasing~\cite{DBLP:conf/emnlp/KrishnaWI20}. Text detoxification is a subtask of text style transfer (TST), which involves rewriting text while preserving its original meaning and altering specific style attribute, such as formality, bias, expressiveness, sentiment, or, in the case of detoxification, toxicity~\cite{DBLP:conf/aaai/FuTPZY18, DBLP:conf/acl/LaiTN20}.

\begin{table}[t!]
    \footnotesize
    \begin{tabular}{l|p{2.79cm}p{2.75cm}}
        \toprule
         & \textbf{Toxic Text} & \textbf{Detoxified Text} \\
        \midrule
        German & Wie be**oppt muss man sein? &  Wie verwirrt muss man sein? \\
        Spanish & Que os den por el c**o. &  Que os dé muy mala suerte. \\
        French & c'est moi at***dé ! je suis tombé ! & C'est moi qui suis tombé ! \\
        Russian & \foreignlanguage{russian}{я мужик а вы г**но} & \foreignlanguage{russian}{Я мужчина, а вы неправы} \\

        \bottomrule
    \end{tabular}
    \caption{Examples of the source toxic texts across different languages and their respective synthetic detoxifications from our \dataset.}
\end{table}

While significant progress has been made in monolingual TST and detoxification, both in supervised and unsupervised settings~\cite{DBLP:conf/emnlp/DaleVDLKSP21, DBLP:conf/acl/LogachevaDUMDKS22, DBLP:conf/emnlp/PourFBVPS23}, multilingual text detoxification remains a largely unsolved problem. This is primarily due to two factors: the scarcity of parallel detoxification data across multiple languages and the suboptimal performance of unsupervised methods in cross-lingual settings~\cite{DBLP:conf/ijcnlp/DementievaMDP23}.

\begin{figure*}
    \centering
    \includegraphics[clip, trim=1.1cm 6.1cm 5cm 6.1cm,width=\linewidth]{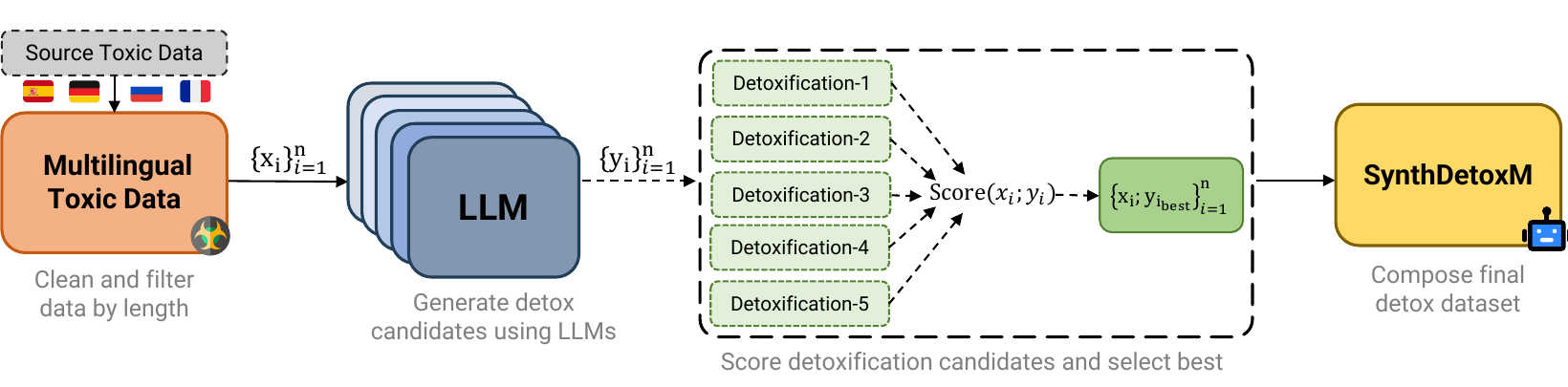}
    \caption{An illustration of the proposed approach for collecting and generating the multilingual text detoxification dataset \dataset.}
    \label{fig:detox-pipeline}
\end{figure*}

Manual or crowdsourced data collection is a challenging and costly task~\cite{DBLP:conf/naacl/RaoT18, reid-artetxe-2023-role, konovalov_collecting_2016}, creating parallel data with the use of modern LLMs, which already proven to work well for the tasks of text classification~\cite{DBLP:conf/emnlp/SunL0WGZ023} and question answering~\cite{DBLP:conf/emnlp/YeGLXF00K22}, remains underexplored. To address these challenges and facilitate the development of multilingual text detoxification models and datasets, we propose a framework for generating parallel multilingual synthetic detoxification data and \dataset, a large-scale multilinugal synthetic parallel text detoxification dataset, which was created using this framework. 

Our dataset is comprised of 16,000 high-quality synthetic detoxification pairs across four languages: German, Spanish, French and Russian. The dataset was created using few-shot prompting and selecting the best generations of five different open-source LLMs. The answers were combined using a handcrafted heuristic, providing the best answers from each model to ensure diversity and quality of the final data.

Our contributions can be summarized as follows:

\begin{enumerate}
    \item We propose a framework for generating synthetic parallel multilingual detoxification data using few-shot prompting of LLMs.
    \item We create \dataset, a large-scale multilingual synthetic parallel dataset for text detoxification, helping to address the data scarcity issue in the detoxification task.
    \item We conduct a thorough empirical evaluation of the proposed dataset, including linguistic analysis of the data and benchmarking against the human-annotated MultiParaDetox.
\end{enumerate}

We openly release the generated data and code.\footnote{\href{https://github.com/s-nlp/synthdetoxm}{github.com/s-nlp/synthdetoxm}} 

\section{Background and Related Work}

\subsection{Text Style Transfer}

Text Style Transfer (TST), the task of rewriting the text in a target style while preserving its semantic content and fluency, has garnered significant attention in the natural language processing community due to its potential applications in text generation~\cite{DBLP:conf/aaai/FuTPZY18}. TST encompasses various subtasks, including formality style transfer~\cite{DBLP:conf/coling/WangWMLC20, DBLP:conf/acl/LaiTN20}, sentiment style transfer~\cite{DBLP:conf/emnlp/YuZLC21}, authorship style transfer~\cite{DBLP:journals/corr/abs-2406-15586, DBLP:journals/corr/abs-2403-08043}, and detoxification~\cite{DBLP:conf/emnlp/DaleVDLKSP21,DBLP:conf/coling/AtwellHA22, DBLP:journals/corr/abs-2206-02252, DBLP:conf/emnlp/MoskovskiyPP24}. 

With the advent of Large Language Models (LLMs), in-context learning methods have increasingly been utilized for TST and detoxification tasks. \citet{DBLP:conf/emnlp/SuzgunMJ22} proposed a novel approach to TST by prompting LLMs and then reranking the generated texts based on three TST metrics: text similarity, target style strength, and fluency. Similarly, \citet{DBLP:conf/acl/ReifIYCCW22} demonstrated the effectiveness of prompting GPT-3, a state-of-the-art LLM at the time, to rewrite texts in a desired style.

\subsection{Text Detoxification}

Text Detoxification, a subtask of Text Style Transfer (TST), involves transforming an input text $x_i$, identified as toxic through toxicity estimation models, into a text $y_i$ that is non-toxic in style while maintaining semantic similarity and fluency. In this context, toxicity refers to language that is harmful, offensive, or inappropriate.


Due to the lack of parallel training data, early research focused on unsupervised detoxification methods~\cite{DBLP:conf/acl/SantosMP18, DBLP:conf/emnlp/DaleVDLKSP21, DBLP:conf/acl/HallinanL0S23, DBLP:conf/emnlp/PourFBVPS23}. For instance,\cite{DBLP:conf/acl/LogachevaDUMDKS22} and APPDIA~\cite{DBLP:conf/coling/AtwellHA22}, has enabled the training of sequence-to-sequence models~\cite{DBLP:conf/acl/LogachevaDUMDKS22, DBLP:conf/emnlp/PourFBVPS23} that outperform most unsupervised approaches in terms of rewritten toxicity, fluency, and semantic similarity. In parallel, \citet{DBLP:conf/emnlp/MoskovskiyPP24} explored the use of activation patching in LLMs to generate synthetic parallel detoxification data for English. Their results demonstrated that training detoxification models on this data yields performance comparable to models trained on manually annotated datasets in automatic evaluations, while achieving superior quality in human assessments.

\subsection{Multilingual Text Style Transfer}

The scarcity of high-quality parallel multilingual detoxification data remains a major challenge in the field. Recently, new non-English parallel datasets have been introduced for various TST tasks, including a Bangla language parallel sentiment style transfer dataset~\cite{mukherjee-etal-2023-low} and the extension of the GYAFC dataset to Portuguese, French, and Italian, resulting in XFORMAL~\cite{DBLP:conf/naacl/BriakouLZT21}. Following the crowdsourcing pipeline introduced by~\cite{DBLP:conf/acl/LogachevaDUMDKS22}, a parallel text detoxification dataset for Russian was collected~\cite{DBLP:journals/corr/abs-2206-02252}. Later, using the similar data annotation pipeline,~\citet{DBLP:conf/clef/DementievaMBAR024} collected 1000 sentence pairs across nine languages, resulting in the MultiParaDetox dataset for a corresponding shared task on multilingual text detoxification. Furthermore, in a more recent work~\citet{DBLP:conf/coling/DementievaBRAR025}, provide an in-depth analysis of toxicity characteristics across languages, exploring descriptive linguistic features that influence detoxification quality.

Nevertheless, the size of MultiParaDetox is far from satisfactory with 1000 sentence pairs per language, only 400 of which are publicly available. The remaining 600 pairs comprised the test set for the multilingual text detoxification shared task~\cite{DBLP:conf/clef/DementievaMBAR024}. Such relatively small dataset may be insufficient for training big multilingual language models for multilingual text detoxification.

To bridge this gap, we present \dataset~- a synthetic parallel detoxification corpus for four European languages, namely, German, Spanish, French, and Russian, with 4000 samples for each language. The dataset creationg pipeline presented in our work can easily be transferred to other languages as well, drastically reducing the cost of annotation for parallel detoxification datasets.

\section{Methodology}

In this section, we describe the pipeline introduced for collecting the multilingual parallel text detoxification dataset, \dataset. A general illustration of our approach is shown in Figure~\ref{fig:detox-pipeline}.  

\subsection{Data Collection}

To create \dataset, we begin by selecting several thousand non-parallel toxic texts from publicly available toxicity identification datasets. We focus on four languages for \dataset: German, French, Spanish and Russian. From these datasets we select only the texts that were marked as toxic by human annotators, excluding non-toxic examples. In cases of multiple annotations, we retained the sample where the majority of annotators classified the sentence as toxic.

To enhance the final data quality, we employ sample-level filtering using the \textbf{STA} and \textbf{SIM} metrics\footnote{Evaluation metrics are described in Section~\cref{sec:sta}} and we also apply data augmentation techniques utilizing the Perspective API~\cite{10.1145/3534678.3539147}. Since the API returns both toxicity scores and toxic spans, we further improve data quality by splitting the source texts into sentences and removing those sentences that do not intersect with the detected toxic spans. This filtering process results in a larger dataset of toxic sentences, and we also split overly long inputs into separate examples.

\begin{figure}
    \centering
    \includegraphics[trim=18 13 15 15, width=0.999\linewidth]{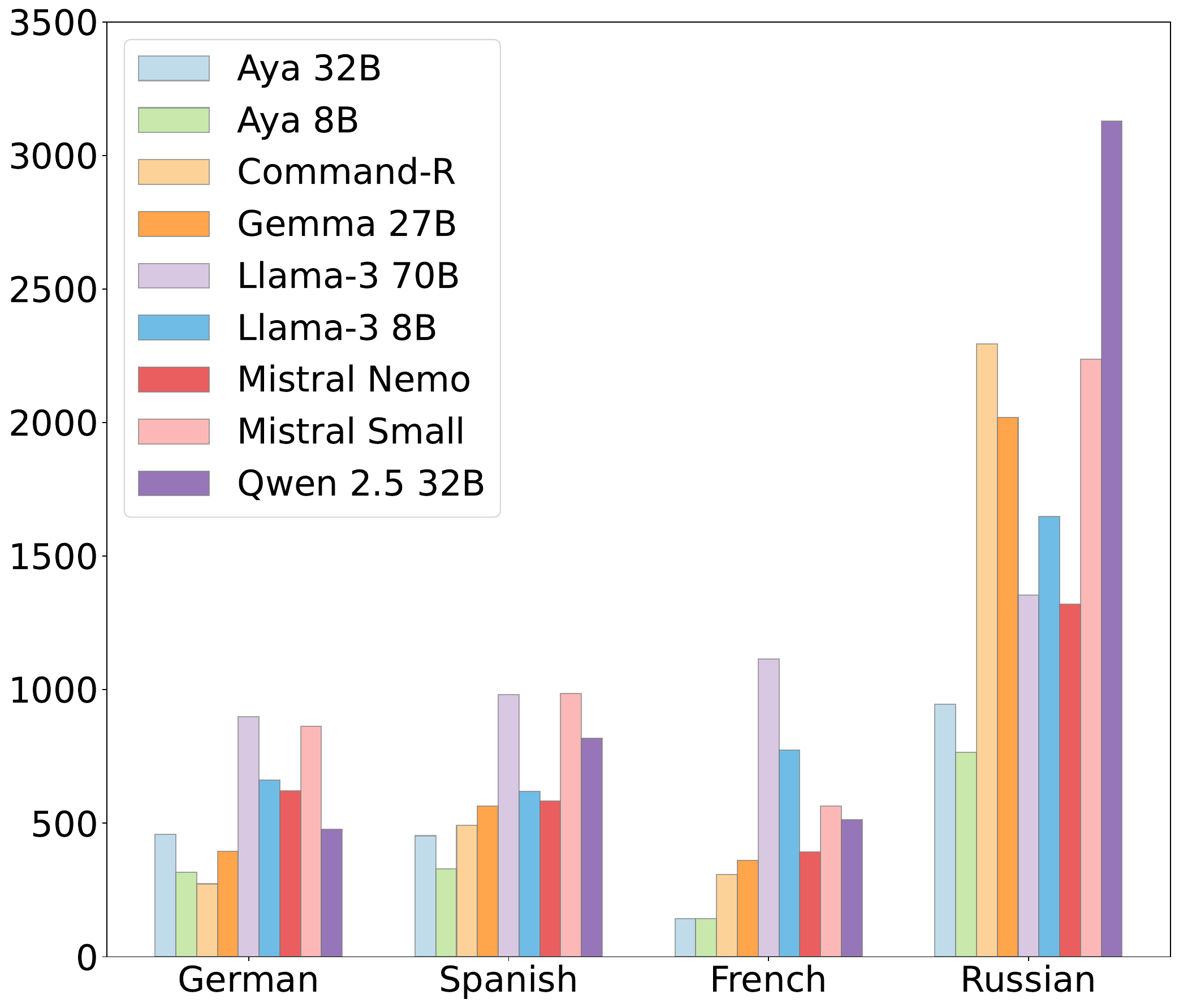}
    \caption{Number of accepted samples in the final \dataset~dataset with respect to the LLM by language.}
    \label{fig:enter-label}
\end{figure}

\paragraph{Russian}

For the Russian dataset, we use data from the Jigsaw Toxic Comments Classification Challenge~\cite{jigsaw-multilingual-toxic-comment-classification}, Russian Language Toxic Comments~\cite{kaggle-russian-language-toxic-comments} and Toxic Russian Comments~\cite{Semiletov2020}. From these sources, we select only those rows labeled as toxic, resulting in more than 15,697 toxic texts. We then calculate the \textbf{STA} and \textbf{SIM} metrics, applying a threshold of 0.5 for filtering. After removing emojis, eliminating texts with fewer than five words or more than 30 words, and splitting the sentences using toxic spans from Perspective API, our final dataset consists of 15,697 texts.

\paragraph{German}  

For German, we use the toxicity identification data from the GermEval 2021 shared task~\cite{risch-etal-2021-overview} and RP-Mod and RP-Crowd~\cite{Assenmacher2021-qk} to create a dataset of 4,946 toxic texts. We apply the same filtering and augmentation pipeline as for the Russian dataset, but with a lower \textbf{STA} score threshold of 0.3. This resulted in a dataset of 4,946 texts, which exceeds the original size of the raw dataset. We attribute this increase to the higher median length of German sentences, which leads to a greater number of split texts.

\paragraph{Spanish}

For Spanish, we utilize data from the Jigsaw Toxic Comments Classification Challenge~\cite{jigsaw-multilingual-toxic-comment-classification} and the Clandestino dataset~\cite{clandestino}. This results in an initial dataset of 10,260 toxic texts. We apply the same filtering and augmentation pipeline as for the German dataset, using the same \textbf{STA} threshold of 0.3. This process yields a final dataset of 5,826 texts.

\paragraph{French}

For French, we use the data from the Jigsaw Toxic Comments Classification Challenge~\cite{jigsaw-multilingual-toxic-comment-classification} and the MLMA Hate Speech Corpus~\cite{ousidhoum-etal-multilingual-hate-speech-2019} to generate a dataset of 5,424 toxic texts. As the toxicity classifier used in other languages does not support French, we instead use the Perspective API to get toxicity scores. After applying a \textbf{STA} score threshold of 0.25, we obtain a dataset of 4,310 sentences.

\subsubsection{Parallel Data Generation Pipeline}

To generate parallel detoxification data, we use various open-source LLMs in a few-shot generation setup. Specifically, we employed the following models: Qwen 2.5 32B by Qwen~\cite{qwen2,qwen2.5}; Command-R 32B by Cohere~\cite{cohere2024}; Gemma 2 27B by Google~\cite{DBLP:journals/corr/abs-2408-00118}; Aya Expanse in 32B and 8B versions by Cohere~\cite{dang2024ayaexpansecombiningresearch}; Mistral Small 22B, Mistral Nemo 12B by Mistral AI~\cite{mistralai2024small, MistralAI2024}; and Llama 3.1 70B and 8B models respectively, by Meta~\cite{grattafiori2024llama3herdmodels}. While not all these models are explicitly designed for multilingual tasks, our experiments show that all of them support the languages considered in this work.

\subsubsection{Few-Shot Example Mining}

To select the best toxic and non-toxic pairs for few-shot generation, we calculate the \textbf{STA} and \textbf{SIM} metrics for all sentences in Russian, German and Spanish from the multilingual toxicity detection dataset\footnote{\href{https://hf.co/datasets/textdetox/multilingual_toxicity_dataset}{hf.co/multilingual\_toxicity\_dataset}}. We then rank the top 10 sentencesbased on the following score:
\begin{align*}
    \boldsymbol{\operatorname{Score}}(x_i; y_i) &=\\ 1 - \left( \frac{1 - \boldsymbol{\operatorname{STA}}(x_i)}{1 - \boldsymbol{\operatorname{STA}}(y_i)} \cdot \left(1 - \boldsymbol{\operatorname{SIM}}(x_i; y_i)\right) \right)
\end{align*}

where $\boldsymbol{\operatorname{STA}}(x_i)$ and $\boldsymbol{\operatorname{STA}}(y_i)$ represent the toxicity scores for the original and detoxified examples, respectively. $\boldsymbol{\operatorname{SIM}}(x_i;y_i)$ is the cosine distance between the embeddings of toxic and detoxified sentences.

This ranking criterion is chosen to ensure high-quality detoxification without altering the original meaning of the sentences. Since the sentences used for few-shot prompting have been annotated by human experts, we expect the detoxification quality to be satisfactory. Additionally, the rewriting process of toxic words often leads to an expanded distance between toxic and non-toxic sentences, increasing the distinction in non-toxicity.  To maximize both the distance and the distinction between the original and detoxified sentences, we select harder and more meaningful examples for few-shot prompting, which helps improve the detoxification process.

For French, which is not represented in the MultiParaDetox dataset, we used human annotators to detoxify 10 randomly chosen sentences from the existing non-parallel data.

After generating detoxified examples, we perform refusal filtering using a refusal classification model~(see details in Appendix~\ref{sec:refusal_clf}). Additionally, we use a simple threshold-based non-detoxifiability metric, calculated by dividing the absolute reduction in the STA score by the original STA score. We compare the resulting detoxifiability scores to a fixed threshold of 0.5. If the score falls below this threshold, the example is considered non-detoxifiable.

After generating five detoxification datasets in each language using the selected models, we rank the sentences by their multiplied \textbf{STA} and \textbf{SIM} metrics and select the top-scoring examples. This metric helps mitigate issues such as refusal(where models refuse to generate text due to toxicity) and copy-paste generation (where the model generates the input toxic sentences without modification), as copy-paste generation typically results in a low \textbf{STA} score, while refusal leads to a low \textbf{SIM} score.

\subsection{Final Composed Dataset}

After all preprocessing, cleaning and filtering steps we compose \dataset~- a manually collected and synthetically paraphrased parallel detoxification dataset on 16,000 toxic and non-toxic text pairs for Spanish, German, Russian and French. 

We show the statistics of detoxification candidate acceptance with respect to each LLM Language-wise in Table~\ref{tab:llm_stats} and Figure~\ref{fig:enter-label}. According to the statistics, Qwen 2.5 generated the most preferrable detoxifications among other models. 

However, upon manual examination we noticed that Qwen tended to occasionally insert tokens of Chinese text into the generated text though was prompted to answer only on the language of the source text. Therefore, the strict reranking and filtering criteria of generated detoxification candidates is necessary.

\subsection{Data Quality Evaluation Pipeline}

To evaluate the quality of our generated detoxification data in Russian, German, and Spanish, we use our dataset for training and compare the performance of models trained on \dataset with those trained on the human-annotated parallel detoxification dataset, MultiParaDetox~\citet{DBLP:conf/clef/DementievaMBAR024}. Due to its absence in the MultiParaDetox dataset, French is excluded from this comparison. A more detailed linguistic analysis of the dataset can be found in Appendix~\ref{sec:linguistic_analysis}.

\begin{figure*}
    \centering
    \includegraphics[trim=40 6 0 11, clip, width=0.66\linewidth]{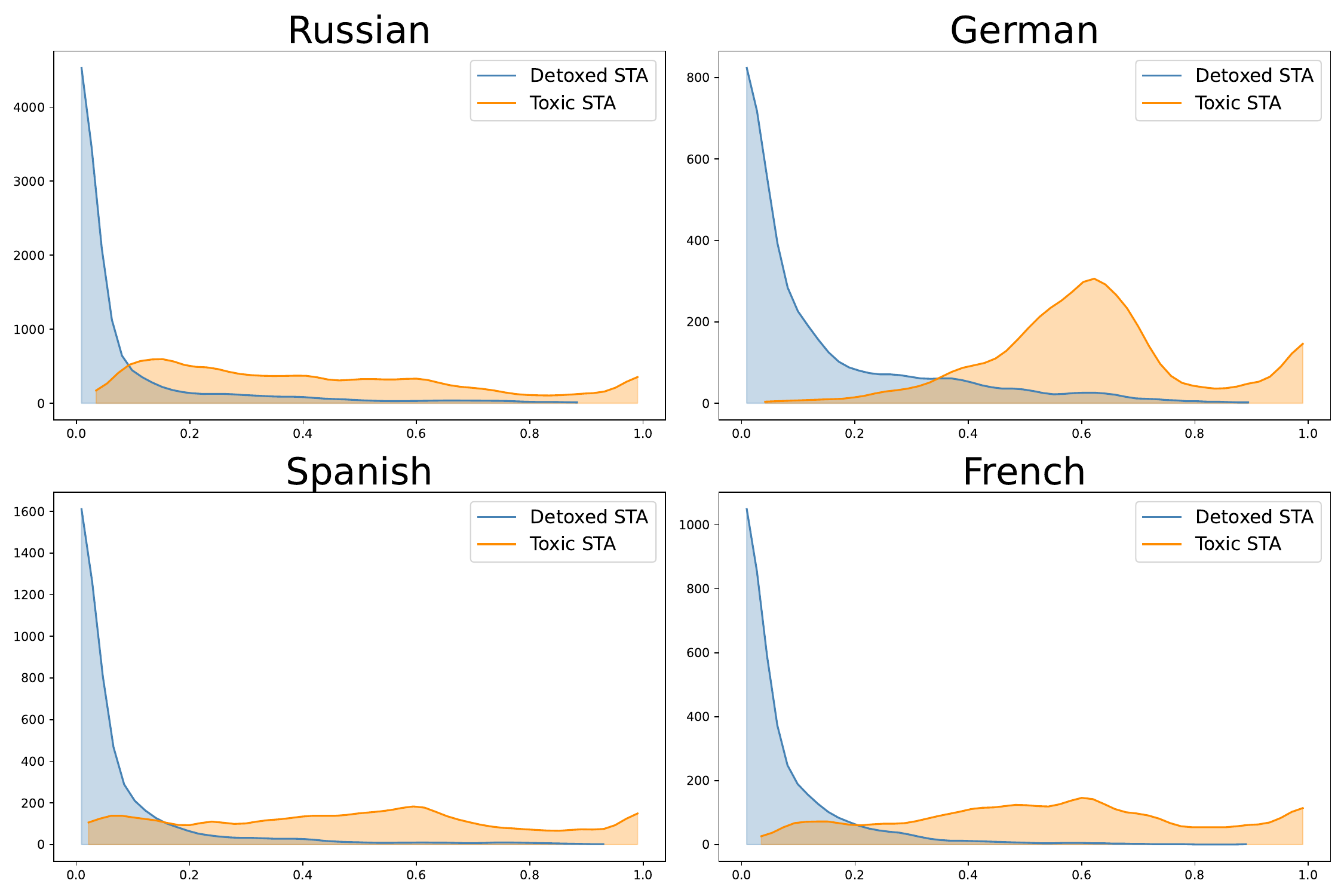}
    \caption{Distribution of STA toxicity scores of toxic and neutral examples in the dataset. The original toxic texts are in orange, while detoxified texts are in blue. For readability we apply Gaussian smoothing.}
    \label{fig:dataset-metrics-figure}
\end{figure*}

\section{Experimental Setup}

\begin{table}[!t]
\centering
\setlength{\tabcolsep}{1.5pt}
\begin{tabular}{lcccc}
\toprule

& \textbf{STA$_\text{T}$}$\uparrow$ & \textbf{STA$_{\text{D}}$}$\uparrow$ & \textbf{SIM}$\uparrow$ & \textbf{STA$_{\text{D}}$}$\times$\textbf{SIM}$\uparrow$ \\ 
\midrule
German & 0.389 & 0.853 & 0.793 & 0.675 \\
Spanish & 0.514 & 0.920 & 0.736 & 0.681 \\
French & 0.583 & 0.913 & 0.677 & 0.624 \\
Russian & 0.467 & 0.924 & 0.731 & 0.678 \\

\bottomrule
\end{tabular}
\caption{Average toxicity levels across different languages for source toxic (T) and generated detoxified (D) texts, along with similarity scores. \textbf{STA$_\text{T}$} represents the toxicity level of the original text, while \textbf{STA$_\text{D}$} corresponds to the detoxified text. In our work, for a text $x$ the score \textbf{STA}$(x)=1-P(\text{toxic}|x)$.}
\label{tab:dataset-metrics}
\end{table}

\subsection{Data Quality Tests}

To evaluate the efficacy of our \dataset~for German, Spanish and Russian, we've trained a series of sequence-to-sequence models on different folds from the dataset. Since MultiParaDetox consists of only 400 pairs of toxic texts with their human-written non-toxic rephrasings, we split our created \dataset~dataset into 10 chunks of 400 pairs for German, Spanish and Russian. We trained 10 mT0 models on different chunks of the dataset and evaluated their average performance on the MultiParaDetox test set. Additionally, we test if using both our \dataset~and MultiParaDetox for training would lead to improved performance. 

\subsection{Toxicity and Similarity of Synthetic Texts}

To further assess the quality of the generated data, we computed the \textbf{STA} and \textbf{SIM} scores using the Perspective API for Russian, German, Spanish, and French. These metrics were selected for their relevance to detoxification tasks and their ability to quantitatively assess our synthetic dataset. We also assessed the quality of the French subset of \dataset, as French is not represented in the MultiParaDetox dataset, and therefore cannot be evaluated through model training. The scores are presented in Figure~\ref{fig:dataset-metrics-figure} and Table~\ref{tab:dataset-metrics}. 

The results indicate that French achieves comparable automatic metric scores to other languages, suggesting that detoxification models trained on this data would perform similarly. Therefore, we hypothesize that the French subset of \dataset~is a valuable addition to the dataset, enabling the training of effective detoxification models for French language processing tasks.

\subsection{Automatic Evaluation Setup}

To assess the quality of the generated Spanish, Russian, and German data, we follow the evaluation pipeline of~\citet{DBLP:conf/clef/DementievaMBAR024}, developed for the multilingual text detoxification shared task. These metrics are inspired by prior work on monolingual text detoxification for English and Russian~\cite{DBLP:conf/acl/LogachevaDUMDKS22,DBLP:journals/corr/abs-2206-02252}.

\paragraph{Style Transfer Accuracy (STA)}
\label{sec:sta}
For computation of this metric we use a multilingual toxicity classifier based on a multilingual XLM-R\footnote{\href{https://hf.co/textdetox/xlmr-large-toxicity-classifier}{hf.co/textdetox/xlmr-large-toxicity-classifier}}~\cite{DBLP:conf/acl/ConneauKGCWGGOZ20} text classification model, trained on a binary toxicity detection dataset.

\paragraph{Content Similarity (SIM)}
\label{sec:sim}
For computation of this metric we use the cosine distance between LaBSE\footnote{\href{https://hf.co/sentence-transformers/LaBSE}{hf.co/sentence-transformers/LaBSE}} embeddings~\cite{DBLP:conf/acl/FengYCA022} of the source texts and the generated texts. 

\paragraph{Fluency (FL)}
Fluency assesses how closely detoxified texts resemble human-written references. Previous works on English have employed CoLA-based classifiers to estimate text fluency~\cite{DBLP:conf/acl/LogachevaDUMDKS22,DBLP:conf/emnlp/MoskovskiyPP24}. However, due to the absence of CoLA datasets for all considered languages~\citet{DBLP:conf/clef/DementievaMBAR024} used ChrF1 as a substitute. While recent work has introduced MELA~\cite{DBLP:conf/acl/ZhangLHMWH24}, a multilingual extension of CoLA covering all the languages in this study, we maintain the evaluation pipeline of \citet{DBLP:conf/clef/DementievaMBAR024} and continue using ChrF1.

Nonetheless, ChrF1 remains a coarse approximation of text fluency, which may negatively impact the overall \textbf{J} scores (see Appendix~\ref{sec:app_chrf} for details).



\paragraph{Joint score (J)} 

The metrics \textbf{STA}, \textbf{SIM} and \textbf{FL} are subsequently combined into the final \textbf{J} score which is used for the final ranking of approaches. Given an input toxic text $x_i$ and its output detoxified version $y_i$, for a test set of $n$ samples:
\begin{center}
    $\textbf{J} = \frac{1}{n}\sum\limits_{i=1}^{n}\textbf{STA}(y_i) \cdot \textbf{SIM}(x_i,y_i) \cdot \textbf{FL}(x_i, y_i)$,
\end{center}
where \textbf{STA}($y_i$), \textbf{SIM}($x_i,y_i$), \textbf{FL}($x_i,y_i$) $\in [0, 1]$ for each text detoxification output $y_i$.

\subsection{Baselines}

In our work we adopt the baselines for multilingual detoxification described in MultiParaDetox~\cite{DBLP:conf/clef/DementievaMBAR024} that are inspired by prior works on text detoxification~\cite{DBLP:conf/acl/LogachevaDUMDKS22, DBLP:conf/ijcnlp/DementievaMDP23}.

\paragraph{Duplicate} is the simplest baseline possible which copies an input toxic sentence. This baseline has $1.0$ (or $100\%$) \textbf{SIM} score by definition. 

\paragraph{Delete} removes the toxic words according to a predefined list of inappropriate words. \citet{DBLP:conf/clef/DementievaMBAR024} collects the lists of such toxic keywords for all target languages based on openly available sources. These lists are available online\footnote{\href{https://hf.co/datasets/textdetox/multilingual_toxic_lexicon}{hf.co/multilingual\_toxic\_lexicon}}.

\paragraph{Backtranslation} has proven to be effective in previous works~\cite{DBLP:conf/ijcnlp/DementievaMDP23, DBLP:conf/acl/TsvetkovBSP18, kon3}. Following~\cite{DBLP:conf/ijcnlp/DementievaMDP23} we translate texts into English with NLLB translation model~\cite{DBLP:journals/corr/abs-2207-04672}\footnote{\href{https://hf.co/facebook/nllb-200-distilled-600M}{hf.co/facebook/nllb-200-distilled-600M}}. The translated data is then detoxified with the ParaDetox BART~\cite{DBLP:conf/acl/LogachevaDUMDKS22} model\footnote{\href{https://hf.co/s-nlp/bart-base-detox}{hf.co/s-nlp/bart-base-detox}}. After that, the detoxified texts are translated back into the source language using NLLB.

\subsection{Training Configuration}

In our experimental evaluation, of the generated multilingual parallel detoxificiation dataset \dataset~we follow the most efficient approaches used during the TextDetox 2024 Shared Task~\cite{DBLP:conf/clef/Sushko24, DBLP:conf/clef/Protasov24, DBLP:conf/clef/RykovZAV24}, where top three solutions in automatic evaluations utilized fine-tuning of a multilingual encoder-decoder language model mT0~\cite{DBLP:conf/acl/MuennighoffWSRB23}. 

We use \texttt{mT0-XL} model\footnote{\href{https://hf.co/bigscience/mt0-xl}{hf.co/bigscience/mt0-xl}} and perform fine-tuning in full precision. We use AdaFactor optimizer~\cite{adafactor} with batch size of $16$, $50$ warmup steps and set maximum sequence length to $512$. 

We fine-tune mT0 for $2$ epochs in all setups. According to our experiments, the increased number of training epochs does not increase the final performance of the model. This might be explained to the overall training data scarcity compared to the size of the model: \texttt{mT0-XL} has 3 billion parameters and is being fine-tuned on 1,200 samples (400 for each of the three languages).

\begin{table}[!t]
\setlength{\tabcolsep}{1.5pt}
\resizebox{0.49\textwidth}{!}{
\begin{tabular}{lccccc}
\toprule
\textbf{Dataset} & \textbf{STA} & \textbf{SIM} & \textbf{FL} & \textbf{J} & \textbf{STA$\cdot$SIM} \\
\midrule
\multicolumn{6}{c}{\textbf{German}} \\
\midrule
MPD & $0.722$ & $0.848$ & $0.602$ & $0.383$ & $0.612$ \\
\textsf{\small SDM}~(Subset) & $0.681$ & $0.912$ & $0.745$ & $0.463$ & $0.597$ \\
\textsf{\small SDM} & $\textbf{0.728}$ & $0.899$ & $0.734$ & $\textbf{0.484}$ & $\textbf{0.655}$ \\
\textsf{\small SDM}+MPD & $0.615$ & $\textbf{0.954}$ & $\textbf{0.821}$ & $0.483$ & $0.586$ \\
\midrule
\multicolumn{6}{c}{\textbf{Russian}} \\
\midrule
MPD & $0.748$ & $0.852$ & $0.643$ & $0.434$ & $0.637$ \\
\textsf{\small SDM}~(Subset) & $0.858$ & $0.850$ & $0.656$ & $0.478$ & $0.729$ \\
\textsf{\small SDM} & $\textbf{0.927}$ & $0.839$ & $0.656$ & $\textbf{0.521}$ & $\textbf{0.778}$ \\
\textsf{\small SDM}+MPD & $0.815$ & $\textbf{0.886}$ & $\textbf{0.726}$ & $0.540$ & $0.721$ \\
\midrule
\multicolumn{6}{c}{\textbf{Spanish}} \\
\midrule
MPD & $0.597$ & $0.880$ & $0.616$ & $0.335$ & $0.525$ \\
\textsf{\small SDM}~(Subset) & $0.795$ & $0.856$ & $0.611$ & $0.416$ & $0.681$ \\
\textsf{\small SDM} & $\textbf{0.864}$ & $0.861$ & $0.621$ & $\textbf{0.471}$ & $\textbf{0.744}$ \\
\textsf{\small SDM}+MPD & $0.681$ & $\textbf{0.907}$ & $\textbf{0.653}$ & $0.413$ & $0.618$ \\
\bottomrule
\end{tabular}}
\caption{Results of the automatic evaluation for \texttt{mT0-XL} on German, Russian, and Spanish trained on original data~(MPD stands for MultiParaDetox), our collected and synthetically generated data~(\textsf{\small SDM} stands for \textsf{SynthDetoxM}) and on their combination~(MultiParaDetox + \textsf{SynthDetoxM}).}
\label{tab:eval_metrics}
\end{table}

\section{Results}

Table~\ref{tab:eval_metrics} presents the results of our experimental evaluation of \dataset. For clarity, the table is divided by language. We compare the performance of \texttt{mT0-XL} trained on human-annotated MultiParaDetox data (MPD) with \texttt{mT0-XL} fine-tuned on two subsets of \dataset: a subset of 400 samples per language, matching the MPD size (denoted as \textsf{\small SDM} (Subset)), and the full \dataset~dataset. Additionally, following prior work~\cite{DBLP:conf/acl/XuWLD0M23}, we investigate whether a two-stage fine-tuning approach—first on \dataset, then on MPD (denoted as \textsf{\small SDM} + MPD)—yields further improvements.

\begin{table}[!t]
    \centering
    \setlength{\tabcolsep}{2pt}
    \begin{tabular}{lccc}
    \toprule
         & \textbf{German} & \textbf{Spanish} & \textbf{Russian} \\
        \midrule
        \rowcolor{Lightgreen}Human References & $0.733$ & $0.709$ & $0.732$ \\
        \midrule
        \multicolumn{4}{c}{\textbf{Baselines}} \\
        \midrule
        \rowcolor{Gray} Duplicate & $0.287$ & $0.090$ & $0.048$ \\
        \rowcolor{Gray} Delete & $0.362$ & $0.319$ & $0.255$ \\
        \rowcolor{Gray} Backtranslation & $0.233$ & $0.275$ & $0.223$ \\
        \midrule
        \multicolumn{4}{c}{\textbf{\texttt{mT0-XL} supervised fine-tuning}} \\
        \midrule
        MultiParaDetox & $0.446$ & $0.344$ & $0.472$ \\
        \textsf{\small SDM}~(Subset)& $0.460$ & $0.402$ & $0.475$ \\
        \textsf{\small SDM} & $\boldsymbol{0.482}$ & $\boldsymbol{0.470}$ & $\boldsymbol{0.546}$ \\
        \midrule
        \multicolumn{4}{c}{\textbf{10-shot LLM prediction}} \\
        \midrule
        Gemma 2 & $0.353$ & $0.380$ & $0.404$ \\ 
        Mistral Nemo & $0.286$ & $0.290$ & $0.258$ \\
        Mistral Small & $0.371$ & $0.308$ & $0.273$ \\
        Command R & $0.328$ & $0.344$ & $0.402$ \\
        Qwen 2.5 & $0.402$ & $0.443$ & $0.428$ \\ 
        Llama 3.1 8B & $0.394$ & $0.341$ & $0.357$ \\ 
        Aya Expanse 8B & $0.305$ & $0.246$ & $0.225$ \\ 
        Aya Expanse 32B & $0.399$ & $0.320$ & $0.323$ \\ 
    \bottomrule
    \end{tabular}
    \caption{Text detoxification results in terms of \textbf{J} scores for German, Spanish, and Russian languages. The best overall results are \text{boldfaced}. The \text{baselines} and \text{human references} are from~\cite{DBLP:conf/clef/DementievaMBAR024}.}
    \label{tab:auto_results_final_filtered}
\end{table}

\begin{figure*}[t!]
    \centering
    \includegraphics[clip, trim=0.3cm 0.2cm 0.2cm 1.9cm, width=0.63\linewidth]{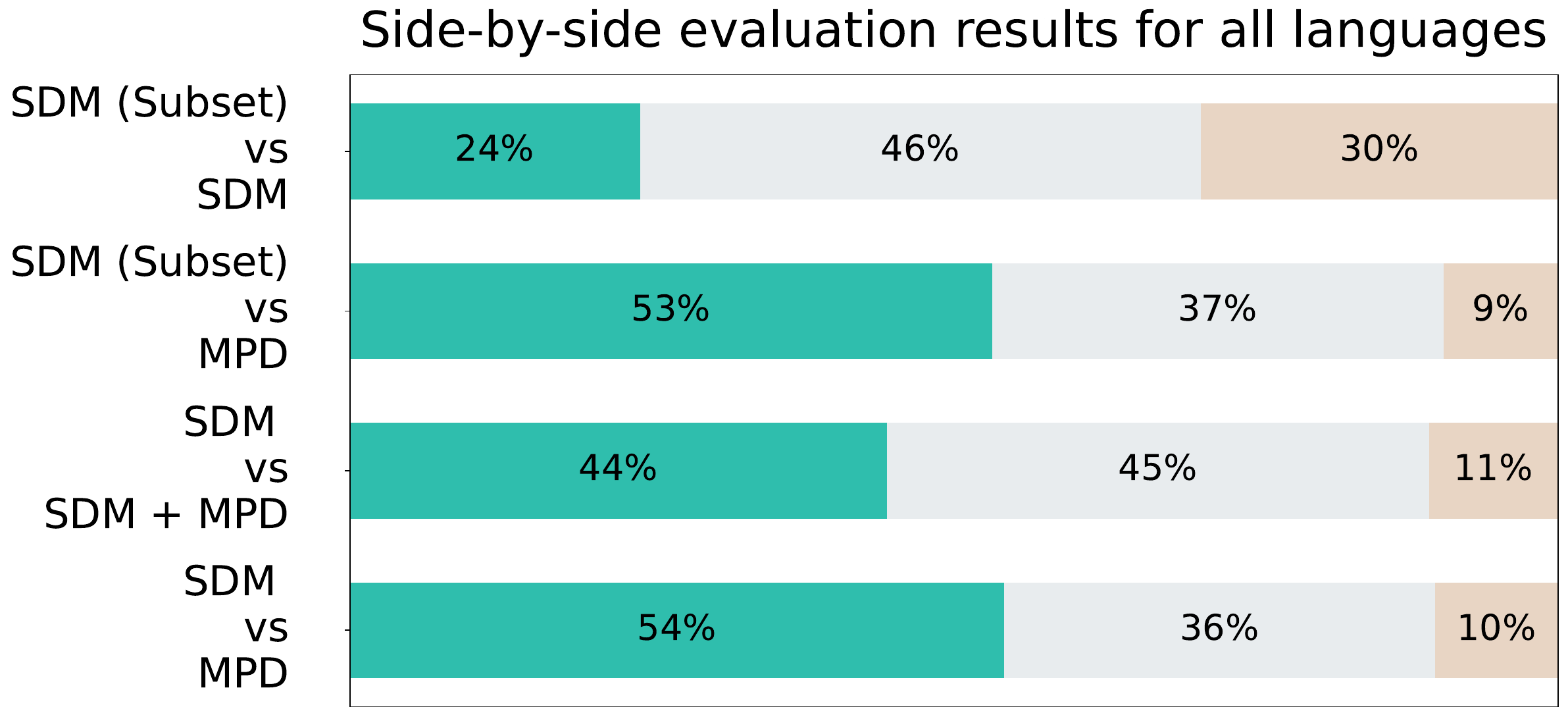}
    \caption{Side-by-side comparison of model outputs across all languages, evaluated by GPT-4o. The results highlight the relative performance of the models in generating detoxified text for German, Russian, and Spanish. The notation is similar to the notation from Table~\ref{tab:eval_metrics}.}
    \label{fig:sbs-overall}
\end{figure*}

The highest \textbf{SIM} scores for German and Russian are achieved by \texttt{mT0-XL} trained on MultiParaDetox (0.954 and 0.886, respectively), with the two-stage training approach (\textsf{\small SDM} + MPD) yielding slightly higher similarity in Russian but lower in German. However, for \textbf{STA}, models trained on \dataset~consistently outperform MultiParaDetox across all languages, both when trained on the full dataset and on a similarly sized subset.

Models trained on \dataset~exhibit slightly lower \textbf{FL} scores, likely due to the reference-dependent nature of this metric. Despite this, the aggregated $\boldsymbol{\operatorname{J}}$ metric—strongly influenced by $\boldsymbol{\operatorname{FL}}$—is significantly higher for models trained on both the full \dataset~and its subset compared to MultiParaDetox. Notably, incorporating MultiParaDetox into the training process (\textsf{\small SDM} + MPD) results in a drop in $\boldsymbol{\operatorname{J}}$ scores.

To further illustrate the advantages of training on \dataset, Table~\ref{tab:eval_metrics} also presents the product of \textbf{STA} and \textbf{SIM}. Even without considering \textbf{FL}, models trained on \dataset~outperform those trained on MultiParaDetox in all setups and languages.

Additionally, Table~\ref{tab:auto_results_final_filtered} reports the $\boldsymbol{\operatorname{J}}$ scores of \texttt{mT0-XL} trained on MultiParaDetox, averaged across 10 subsets of \dataset~and the full \dataset, compared to baselines and large language model (LLM)-based detoxification in a 10-shot generation setting.

Finally, we provide a Side-by-Side (SBS) comparison of the fine-tuned \texttt{mT0-XL} models to evaluate their detoxification performance. Following~\cite{DBLP:conf/emnlp/MoskovskiyPP24}, we employ GPT-4o as an evaluator to select the preferred detoxified outputs. The results of this comparison across German, Spanish, and Russian languages are summarized in Figure~\ref{fig:sbs-overall} and detailed in Appendix~\ref{sec:sbs_mt0}.

Our SBS evaluation shows a clear preference for detoxifications produced by \textsf{\small SDM} over MultiParaDetox (MPD), with \textsf{\small SDM} winning in 59\% of cases compared to 19\% for MPD, and 22\% resulting in ties. The subset of \textsf{\small SDM} also outperforms MPD in 58\% of cases.

When comparing \textsf{\small SDM} against its combination with MPD (\textsf{\small SDM} + MPD), \textsf{\small SDM} is preferred in 47\% of cases, with 21\% favoring \textsf{\small SDM} + MPD, and 33\% tied. Additionally, the full \textsf{\small SDM} dataset is slightly preferred over the batch processing version in 55\% of cases, with 35\% ties. See Appendix~\ref{sec:sbs_mt0} for more details.

\section{Conclusions}

We present  several contributions to multilingual text detoxification technology. Firstly, we successfully extend the concept of few-shot prompting for detoxification to a multilingual context, building upon previous monolingual approaches and propose a framework for generation of multilingual synthetic detoxification data. Secondly, we introduce \dataset, a large-scale multilingual synthetic parallel dataset designed to address the long-standing issue of data scarcity in text detoxification research. Notably, our dataset, created using our selection criteria, demonstrates competitive quality to existing human-annotated datasets, surpassing them in both low resource and high resource settings.

Our comprehensive evaluation of \dataset~reveals its effectiveness in training high-performing models for text detoxification. Specifically, our experiments show that models trained on our dataset \text{outperform} those, which were trained on a similar amount of human-annotated data. Furthermore, training a detoxification encoder-decoder model on full \dataset~yields a model, which surpasses the performance of most large language models in few-shot generation setups. 

Findings presented in our work, show usefulness of the generated data for the task of multilingual text detoxification and pave the way for future research and developments of related technologies.

\section*{Acknowledgments}
The contribution of E.T. was supported by a grant for research centers in the field of artificial intelligence, provided by the Analytical Center for the Government of the Russian Federation in accordance with the subsidy agreement (agreement identifier 000000D730321P5Q0002) and the agreement with the Ivannikov Institute for System Programming of the Russian Academy of Sciences dated November 2, 2021 No. 70-2021-00142. 

\section{Limitations}

One of the limitations of our work is that we are focusing only on explicit type of toxicity. Additionally, definition and type of toxicity changes drastically between the language, e.g. things, that are toxic in one language may be perfectly normal in other language.

Another limitation of this work is our constraint with computational resources, which led to our use of smaller and simpler models for synthetic data generation, which could fit into a single NVIDIA A100 80GB GPU. Usage of larger could potentially result in higher quality and diversity of synthetic data.

Moreover, the comparison with proprietary models would strengthen the evaluation as it is done in recent works~\citet{DBLP:conf/coling/DementievaBRAR025}.

Additionally, we were limited by the amount of annotated non-parallel toxic datasets in some of the languages, which limited the amount of possible generated synthetic data. In future, we plan to extend our work to other languages, such as Italian, Polish and others.

\section{Ethical Considerations}

While working with the task detoxification we are fully aware of the ethical responsibilities involved. As researchers, we handle this sensitive area with care and integrity. The main goal of text detoxification is to make online interactions safer and more inclusive by reducing harmful or offensive language.

While these datasets are meant to train models to detect and reduce toxic language, there's a chance they could be used in the wrong way—such as creating models that spread harmful or offensive content. This could lead to hate speech and harassment.

It's also important to clarify that the goal of text detoxification isn't to suppress free speech or force automatic changes to content. Instead, we aim to build models that offer non-toxic alternatives, helping users choose better language on their own. By giving suggestions rather than enforcing edits, we respect people's freedom while encouraging a more positive online environment.

\bibliography{custom}

\appendix

\section{Prompts}
\subsection{LLM Detoxification Prompt}
\label{sec:detox_prompt}

In our work, for all of the language models we use the Text Style Transfer prompt taken from work by~\citet{DBLP:journals/corr/abs-2402-15951}. Among others, this prompt shown to be giving better generation quality and helped to avoid additional content.

\begin{figure}[htbp!]
    \centering
    \tcbset{colback=grey!5!white, colframe=yellow!75!gray, width=\columnwidth, arc=0mm, auto outer arc, boxrule=0.15mm}
    \begin{tcolorbox}
        Your task is text style transfer. Rewrite the following text into non-toxic language. You must match the target style and preserve the original meaning as much as possible. Do not include the input text in the response. Only generate the target text and nothing else. Make sure the response is completely non-toxic and neutral. Answer only in \{\texttt{language}\}. Here are few examples:",
        Toxic text: \textcolor{red}{\{\texttt{toxic\_text}\}}. Neutral text:
    \end{tcolorbox}
    \caption{Detoxification prompt we use for synthetic parallel data generation. \textcolor{red}{\{\texttt{toxic\_text}\}} stands for a placeholder for a given toxic text being prompted into LLM. In few-shot setting we add few examples of detoxification before last two lines and write: \textit{Here are few examples:}.}
    \label{fig:detox_prompt}
\end{figure}

We slightly adjust it specifically to text detoxification and provide the full text of the prompt in Figure~\ref{fig:detox_prompt}. In this work, we do not use any advanced prompting techniques except few-shot prompting. 

\subsection{mT0 Prompt}

\begin{table}[htbp!]
\centering
\setlength{\tabcolsep}{1pt}
\begin{tabular}{lcccc}
\toprule
Model & German & Spanish & French & Russian \\
\midrule
Llama 3.1 8B & 662 & 619 & 773 & 1648 \\
Llama 3.1 70B & 898 & 981 & 1114 & 1354 \\
Mistral Nemo & 622 & 583 & 392 & 1320 \\
Mistral Small & 862 & 985 & 565 & 2237 \\
Qwen 2.5 32B & 477 & 819 & 513 & 3128 \\
Aya Exp. 32B & 458 & 453 & 142 & 945 \\
Aya Exp. 8B & 316 & 330 & 143 & 765 \\
Command-R 32B & 273 & 492 & 308 & 2294 \\
Gemma 2 27B & 394 & 564 & 360 & 2019 \\
\bottomrule
\end{tabular}
\caption{Number of accepted samples in the final \dataset~dataset, broken down by language and LLMs.}
\label{tab:llm_stats}
\end{table}

\begin{figure}[htbp!]
    \centering
    \tcbset{colback=grey!5!white, colframe=yellow!75!gray, width=\columnwidth, arc=0mm, auto outer arc, boxrule=0.15mm}
    \begin{tcolorbox}
        Write a non-toxic version of the following text in \{\texttt{language}\}: \{\texttt{toxic\_text}\}
    \end{tcolorbox}
    \caption{Detoxification prompt we use for mT0.}
    \label{fig:detox_prompt_mt0}
\end{figure}

We add specific prompt to mT0 during both training and predictions. In our work do not translate the prompt into the language of generation. Instead, we follow the promping approach from the original paper~\cite{DBLP:conf/acl/MuennighoffWSRB23}. We show the prompt in Figure~\ref{fig:detox_prompt_mt0}.

\section{Automatic evaluation results}

The automatic evaluation results are presented in Table \ref{tab:eval_metrics_appendix}.

\begin{table}[htbp!]
\setlength{\tabcolsep}{1.5pt}
\resizebox{0.49\textwidth}{!}{
\begin{tabular}{lcccc}
\toprule
\textbf{Dataset} & \textbf{STA} & \textbf{SIM} & \textbf{CHRF} & \textbf{J} \\
\midrule
\multicolumn{5}{c}{\textbf{German}} \\
\midrule
MPD & $0.722$ & $0.848$ & $0.602$ & $0.383$ \\
\textsf{\small SDM}~(Subset) & $0.681$ \scalebox{0.7}{$\pm 0.213$} & $0.912$ \scalebox{0.7}{$\pm 0.042$} & $0.745$ \scalebox{0.7}{$\pm 0.035$} & $0.463$ \scalebox{0.7}{$\pm 0.117$} \\
\textsf{\small SDM}~(Full) & $0.728$ & $0.899$ & $0.734$ & $0.484$ \\
SDM+MPD & $0.615$ & $0.954$ & $0.821$ & $0.483$ \\
\midrule
\multicolumn{5}{c}{\textbf{Russian}} \\
\midrule
MPD & $0.748$ & $0.852$ & $0.643$ & $0.434$ \\
\textsf{\small SDM}~(Subset) & $0.858$ \scalebox{0.7}{$\pm 0.034$} & $0.850$ \scalebox{0.7}{$\pm 0.020$} & $0.656$ \scalebox{0.7}{$\pm 0.021$} & $0.478$ \scalebox{0.7}{$\pm 0.014$} \\
\textsf{\small SDM}~(Full) & $0.927$ & $0.839$ & $0.656$ & $0.521$ \\
SDM+MPD & $0.815$ & $0.886$ & $0.726$ & $0.540$ \\
\midrule
\multicolumn{5}{c}{\textbf{Spanish}} \\
\midrule
MPD & $0.597$ & $0.880$ & $0.616$ & $0.335$ \\
\textsf{\small SDM}~(Subset) & $0.795$ \scalebox{0.7}{$\pm 0.083$} & $0.856$ \scalebox{0.7}{$\pm 0.031$} & $0.611$ \scalebox{0.7}{$\pm 0.022$} & $0.416$ \scalebox{0.7}{$\pm 0.023$} \\
\textsf{\small SDM}~(Full) & $0.864$ & $0.861$ & $0.621$ & $0.471$ \\
SDM+MPD & $0.681$ & $0.907$ & $0.653$ & $0.413$ \\
\bottomrule
\end{tabular}}
\caption{Results of the automatic evaluation for \texttt{mT0-XL} on German, Russian, and Spanish trained on original data~(MPD stands for MultiParaDetox), our collected and synthetically generated data~(\textsf{\small SDM} stands for \dataset) and on their combination~(MultiParaDetox + \dataset).}
\label{tab:eval_metrics_appendix}
\end{table}

\section{Limitations of ChrF1 as a Fluency Metric}
\label{sec:app_chrf}

This section addresses the issues with using ChrF1 to evaluate fluency in text detoxification. While ChrF1 is commonly used in neural machine translation~\cite{DBLP:conf/wmt/Popovic15}, it has significant drawbacks for text style transfer tasks like detoxification.

Reference-based metrics like ChrF1 are ill-suited for assessing fluency in detoxification. The goal is to change the text's style while maintaining meaning and fluency, without limiting the extent of edits. Effective detoxification often involves substantial structural changes, making comparisons with the original toxic text using ChrF1 misleading. Though ChrF1 may produce low scores, manual evaluations frequently show that the detoxified output is fluent.

ChrF1, based on character n-grams, is sensitive to word order and structural changes, which are often necessary for detoxification. It also fails to consider semantic content, meaning fluency can be high even when the ChrF1 score is low. Additionally, it tends to reward minimal edits, which undermines the goal of thorough detoxification.

\begin{table}[!t]
    \centering
    \begin{tabular}{lccc}
        \toprule
        & Spanish$\downarrow$ & German$\downarrow$ & Russian$\downarrow$ \\
        \midrule
        Toxic & 2089 & 323 & 4467 \\
        Detoxified & 27 & 102 & 14 \\
        \bottomrule
    \end{tabular}
    \caption{Total amount of toxic words for toxic and detoxified subsets of \dataset~with respect to language.}
    \label{tab:word_counts}
\end{table}

\begin{table}[!t]
    \centering
    \begin{tabular}{lccc}
        \toprule
        & Spanish$\downarrow$ & German$\downarrow$ & Russian$\downarrow$ \\
        \midrule
        Toxic & 0.522 & 0.081 & 1.117 \\
        Detoxified & 0.007 & 0.036 & 0.004 \\
        \bottomrule
    \end{tabular}
    \caption{Average number of toxic words per text in the toxic and detoxified \dataset~with respect to language.}
    \label{tab:mean_toxic_words}
\end{table}

Recent research has shifted towards more appropriate fluency metrics. For example, CoLA-based classifiers, as used in \citet{DBLP:conf/ijcnlp/DementievaMDP23} and \citet{DBLP:conf/acl/LogachevaDUMDKS22}, focus on linguistic acceptability, offering a more accurate assessment of fluency without relying on comparisons to the toxic input.

While ChrF1 has its merits in other tasks, it is not suitable for evaluating fluency in detoxification. Future work should prioritize methods that assess fluency based on grammaticality and naturalness, independent of the original text.

\section{Refusal classifier training}
\label{sec:refusal_clf}

\begin{figure}[t!]
    \centering
    \tcbset{colback=grey!5!white, colframe=yellow!75!gray, width=\columnwidth, arc=0mm, auto outer arc, boxrule=0.15mm}
    \begin{tcolorbox}
        Politely refuse to answer this in \texttt{\{lang\}} and provide an explanation why you refuse. The refusal should be connected to the request topic. Do not add anything additional, only respond with a refusal: \texttt{\{input\_text\}}
    \end{tcolorbox}
    \caption{Refusal generation prompt for synthetic refusals dataset.}
    \label{fig:refusal_synthetic_prompt}
\end{figure}

To get rid of LLM refusals in \dataset, we trained a separate refusal classifier, based on \texttt{xlmr-base}\footnote{\href{https://hf.co/s-nlp/xlmr-base-refusal-classifier}{hf.co/s-nlp/xlmr-base-refusal-classifier}}. 

To train the model, a high quality synthetic dataset was created\footnote{\href{https://hf.co/datasets/chameleon-lizard/multilingual_refusals}{hf.co/datasets/chameleon-lizard/multilingual\_refusals}}. It was based on randomly selected inputs from the LMSYS-Chat-1M~\cite{zheng2023lmsyschat1m} dataset and then passed to the Gemini Flash 1.5~\cite{geminiteam2024gemini15unlockingmultimodal} and Llama 3.3 70B~\cite{grattafiori2024llama3herdmodels} models, prompted to generate both responses to the prompts from the dataset and refusals. Prompt for synthetic refusal generation is presented in Figure \ref{fig:refusal_synthetic_prompt}.

Classification model was trained using a batch size of 64, learning rate of 1e-4 for one epoch.

\begin{figure}[t!]
    \centering
    \includegraphics[clip, trim=0.3cm 0.2cm 0.2cm 0.2cm, width=1\linewidth]{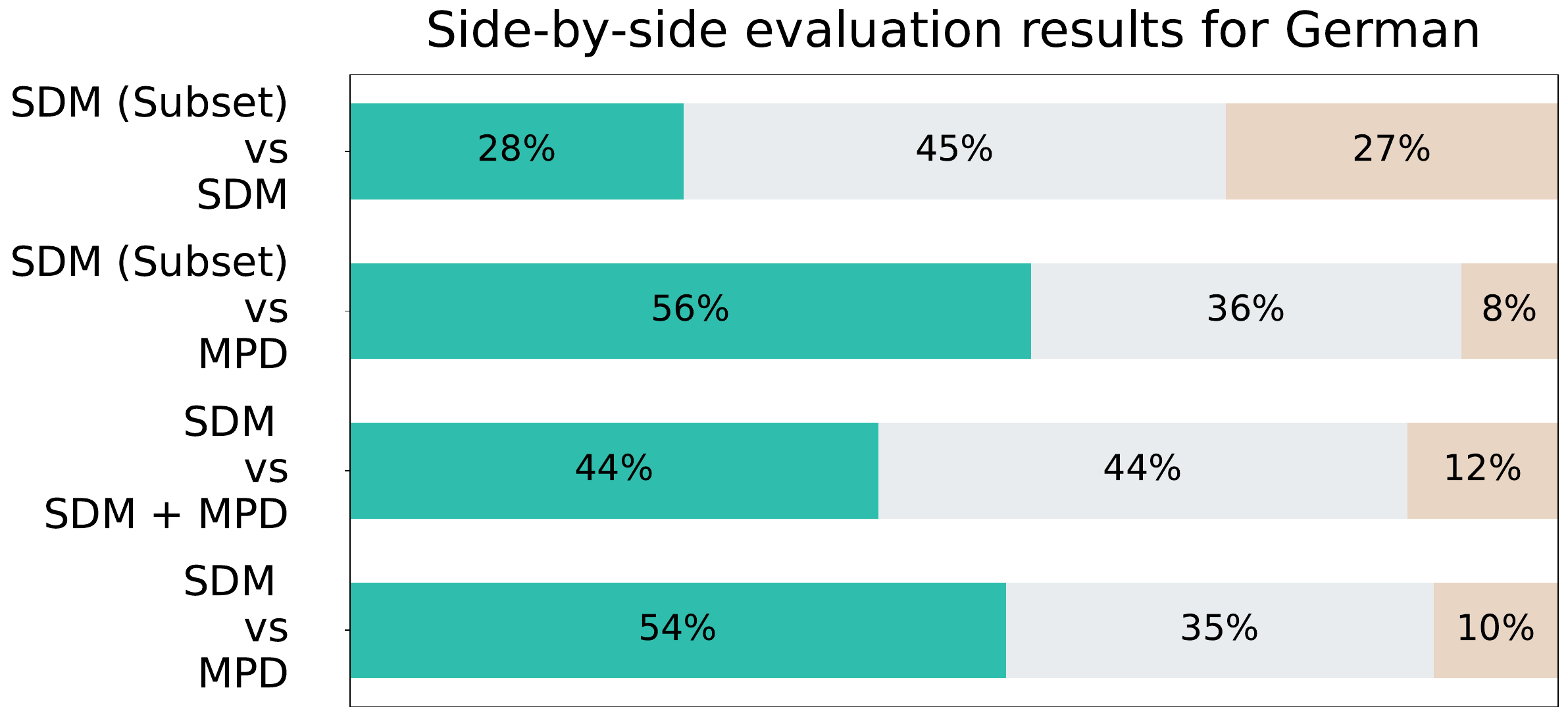}
    \caption{Side-by-side comparison of model outputs across all languages, evaluated by GPT-4o. The results highlight the relative performance of the models in generating detoxified text for German. The notation is similar to the notation from Table~\ref{tab:eval_metrics}.}
    \label{fig:sbs-de}
\end{figure}

\begin{figure}[t!]
    \centering
    \includegraphics[clip, trim=0.3cm 0.2cm 0.2cm 0.2cm, width=1\linewidth]{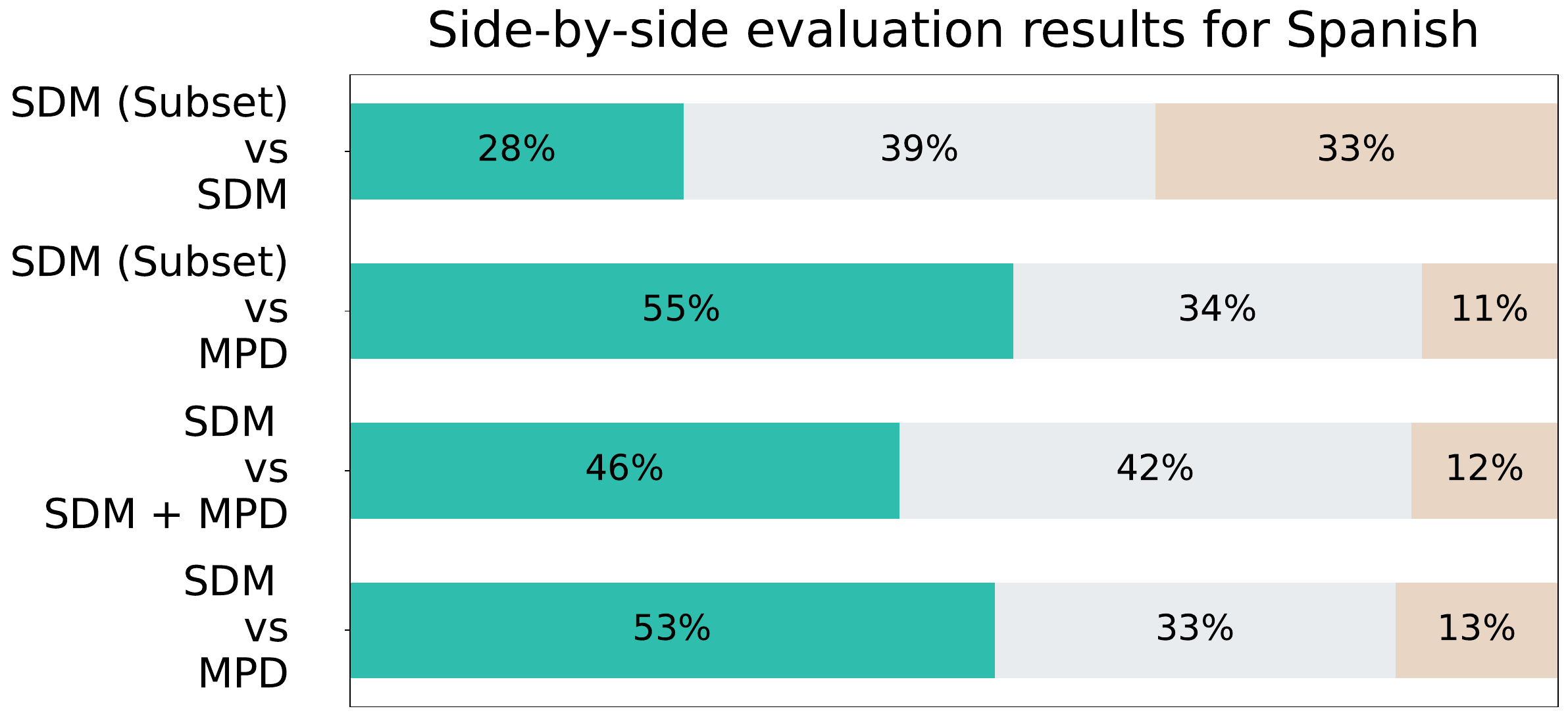}
    \caption{Side-by-side comparison of model outputs across all languages, evaluated by GPT-4o. The results highlight the relative performance of the models in generating detoxified text for Spanish. The notation is similar to the notation from Table~\ref{tab:eval_metrics}.}
    \label{fig:sbs-es}
\end{figure}

\begin{figure}[t!]
    \centering
    \includegraphics[clip, trim=0.3cm 0.2cm 0.2cm 0.2cm, width=1\linewidth]{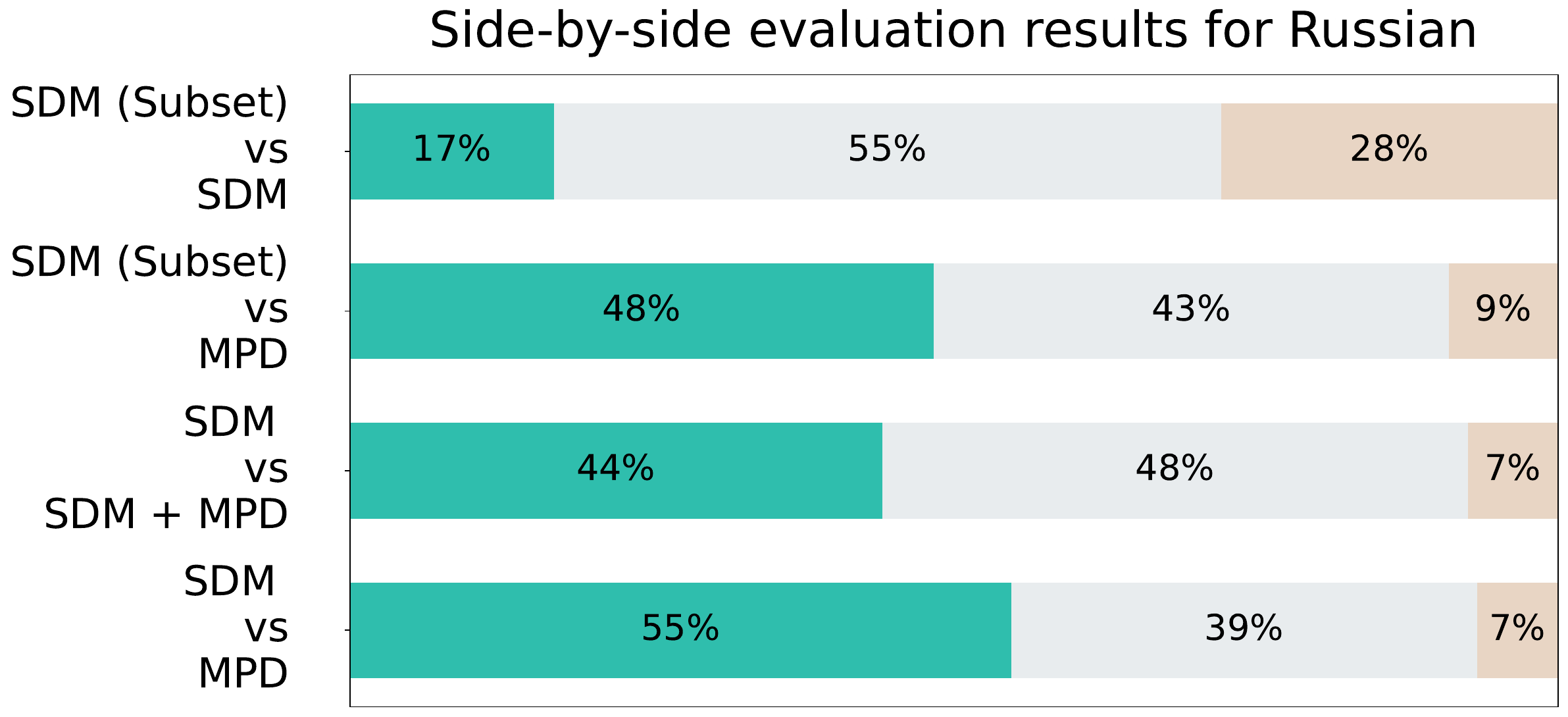}
    \caption{Side-by-side comparison of model outputs across all languages, evaluated by GPT-4o. The results highlight the relative performance of the models in generating detoxified text for Russian. The notation is similar to the notation from Table~\ref{tab:eval_metrics}.}
    \label{fig:sbs-ru}
\end{figure}

\section{Additional linguistic analysis of the dataset}
\label{sec:linguistic_analysis}

To provide additional perspective about detoxification quality of our dataset, we used dataset\footnote{\href{https://hf.co/datasets/textdetox/multilingual_toxic_lexicon}{hf.co/datasets/textdetox/multilingual\_toxic\_lexicon}}, which contains toxic lexicon in German, Spanish and Russian languages and calculated two metrics of our generated data: the amount and mean counts of toxic words in each sentence (presented in Tables \ref{tab:word_counts} and \ref{tab:mean_toxic_words} accordingly) in the toxic and generated detoxified subsets of \dataset.


Due to the lack of French toxic lexicon dataset we did not do any evaluations in French. Furthermore, selected dataset is not comprehensive: for instance, German subset contains only 247 toxic words, which leaves some toxic sentences not having any toxic words detected and overall toxicity of German subset of our dataset is comparatively low. However, these evaluations still show that our detoxified data contains substantially less toxic lexicon than the original toxic data, indicating that overall explicit toxicity of the generated data is much lower after the detoxification.

\section{Side-by-side comparison of trained mT0 models}
\label{sec:sbs_mt0}

To provide additional validation outside of methodology created by \cite{DBLP:conf/clef/DementievaMBAR024}, we have evaluated the responses from the test dataset of each of the trained mT0 models using GPT-4o~\cite{openai2024gpt4ocard} as a judge. To lessen the positional bias, each side-by-side comparison was done twice, changing the positions of the answers and calculating mean score for all answers. The results are presented in the Figures \ref{fig:sbs-de}, \ref{fig:sbs-es}, \ref{fig:sbs-ru}.

As demonstrated in the figures, models fine-tuned on \dataset~achieve significantly higher win rates compared to those trained on the "gold-standard" MultiParaDetox dataset, as well as models trained using a two-stage regime. We attribute this superior performance to the challenges posed by human-sourced samples, which may be too complex or nuanced for relatively simple and compact student models to learn effectively. This limitation likely contributes to the weaker performance of models trained on MultiParaDetox in both automatic evaluations and side-by-side comparisons.

\onecolumn
\section{Per-Language Few-Shot Examples}
\label{app:few-shot-examples}

To ensure reproducibility we show all the few-shot prompts \textit{as is} in Tables~\ref{tab:few_shots_rus}~\ref{tab:few_shots_de},~\ref{tab:few_shots_es}, and~\ref{tab:few_shots_fr}.

\begin{table*}[h!]
    \footnotesize
    \renewcommand{\arraystretch}{1.2}
    \begin{tabular}{l|p{9cm}|c|c}
    \toprule
        Type & Text & Perspective API Score & XLM-R Score \\
        \midrule
        Toxic & \foreignlanguage{russian}{тварь присягу один раз,а проебал всю страну} \newline \textit{\scriptsize The bastard swore an oath once, but screwed over the whole country.} & 0.820 & 1.000 \\
        Neutral & \foreignlanguage{russian}{Присягу дал один раз, а разрушил всю страну} \newline \textit{\scriptsize He took the oath once, but destroyed the whole country} & 0.360 & 0.001 \\
        \midrule
        Toxic & \foreignlanguage{russian}{нахрена этого придурка клоуна сюда впихивают?} \newline \textit{\scriptsize why the fuck is this idiot clown being pushed in here?} & 0.743 & 1.000 \\
        Neutral & \foreignlanguage{russian}{Зачем этого человека сюда впихивают?} \newline \textit{\scriptsize Why is this man being pushed in here?} & 0.042 & 0.005 \\
        \midrule
        Toxic & \foreignlanguage{russian}{молись не молись,всё - равно получишь хуй. ну, разве что блядям на радость} \newline \textit{\scriptsize Pray or don't pray, you'll get a dick anyway. Well, except for the joy of whores.} & 0.757 & 1.000 \\
        Neutral & \foreignlanguage{russian}{Молись не молись, всё - равно ничего не получишь. ну, разве что завистникам на радость.} \newline \textit{\scriptsize Pray or don't pray, you won't get anything anyway. well, except for the envy of the joy.} & 0.156 & 0.998 \\
        \midrule
        Toxic & \foreignlanguage{russian}{её муженька козла на кол надо посадить} \newline \textit{\scriptsize Her scoundrel of a husband needs to be impaled.} & 0.553 & 0.999 \\
        Neutral & \foreignlanguage{russian}{Её муженька нужно сурово наказать} \newline \textit{\scriptsize Her hubby needs to be severely punished} & 0.029 & 0.988 \\
        \midrule
        Toxic & \foreignlanguage{russian}{хорошо пизды не получил уёбок} \newline \textit{\scriptsize good thing the bastard didn’t get his ass kicked.} & 0.853 & 1.000 \\
        Neutral & \foreignlanguage{russian}{хорошо что он не был избит} \newline \textit{\scriptsize it's a good thing he wasn't beaten up} & 0.037 & 0.000 \\
        \midrule
        Toxic & \foreignlanguage{russian}{да что ты доебалась до этих клумб? людям жрать нечего! по мусоркам лазают!} \newline \textit{\scriptsize why are you so fixated on those flower beds? people have nothing to eat! they’re rummaging through trash!} & 0.548 & 1.000 \\
        Neutral & \foreignlanguage{russian}{да что ты докопалась до этих клумб? Людям есть нечего! по мусоркам лазают!} \newline \textit{\scriptsize why did you get to these flower beds? People have nothing to eat! they're climbing through the trash!} & 0.115 & 0.000 \\
        \midrule
        Toxic & \foreignlanguage{russian}{именно благодаря этому президенту, россия еще жива. а такие долбоёбы, только могут воду мутить. дебилоид.} \newline \textit{\scriptsize It's precisely thanks to this president that Russia is still alive. And idiots like you can only stir up trouble. Moron.} & 0.863 & 1.000 \\
        Neutral & \foreignlanguage{russian}{Именно благодаря этому президенту, Россия ещё жива, а такие люди, как ты, только и могут, что наводить смуту.} \newline \textit{\scriptsize It is thanks to this president that Russia is still alive, and people like you can only cause trouble.} & 0.369 & 0.003 \\
        \midrule
        Toxic & \foreignlanguage{russian}{ты просто пидор,а не немец.я за свои слова отвечу,} \newline \textit{\scriptsize you're just a faggot, not a german. I stand by my words, but you won’t, because you're a couch scumbag.} & 0.906 & 1.000 \\
        Neutral & \foreignlanguage{russian}{ты просто не немец. я за свои слова отвечу, а ты н} \newline \textit{\scriptsize you're just not german. i'll answer for my words, but you won't, because you're writing from the couch} & 0.399 & 0.002 \\
        \midrule
        Toxic & \foreignlanguage{russian}{да, блядь, кто это говнище постоянно сюда сливает?!} \newline \textit{\scriptsize Yeah, fuck, who keeps dumping this shit here all the time?!} & 0.654 & 1.000 \\
        Neutral & \foreignlanguage{russian}{да кто эту ерунду постоянно сюда сливает?!} \newline \textit{\scriptsize who is constantly pouring this nonsense here?!} & 0.096 & 0.001 \\
        \midrule
        Toxic & \foreignlanguage{russian}{да вы заебали этим постом уже, пидарасы!!!} \newline \textit{\scriptsize I'm sick of this post already, you bastards!!!} & 0.903 & 1.000 \\
        Neutral & \foreignlanguage{russian}{да вы надоели этим постом уже, люди!!!} \newline \textit{\scriptsize Yes, people, I am tired of this post already!!!} & 0.012 & 0.001 \\
        \bottomrule
    \end{tabular}
    \caption{Selected few-shot examples for the Russian language with toxicity scores.}
    \label{tab:few_shots_rus}
\end{table*}

\begin{table*}[h!]
    \centering
    \footnotesize
    \renewcommand{\arraystretch}{1.2}
    \scalebox{0.95}{
    \begin{tabular}{l|p{9cm}|c|c}
    \toprule
        Type & Text & Perspective API Score & XLM-R Score \\
        \midrule
        Toxic & \foreignlanguage{german}{In dem Schwarz Grün regierden Hessen werden jetzt 8 zusätzliche Psychologen nur für die Jugendlichen Invasoren eingestellt. Jetzt AFD wählen} \newline \textit{\scriptsize In the black-green governed Hesse, 8 additional psychologists are now being hired only for the young invader. Vote AFD now} & 0.397 & 0.486 \\
        Neutral & \foreignlanguage{german}{In dem Schwarz Grün regierden Hessen werden jetzt 8 zusätzliche Psychologen nur für die Jugendlichen Einwanderer eingestellt. Jetzt AFD wählen} \newline \textit{\scriptsize In the black-green governed Hesse, 8 additional psychologists are now being hired only for the young immigrants. Vote AFD now} & 0.313 & 0.004 \\
        \midrule
        Toxic & \foreignlanguage{german}{Moin, moin, seid Ihr alle noch gesund und munter bei all dem geistigen Dünnpfiff im "Merkel-Propagandaschrott"?} \newline \textit{\scriptsize Hello, hello, are you all still healthy and cheerful despite all the intellectual drivel in the "Merkel propaganda junk"?} & 0.471 & 0.898 \\
        Neutral & \foreignlanguage{german}{Moin, moin, seid Ihr alle noch gesund und munter bei all dem geistigen Quatsch in der "Merkel-Propaganda"?} \newline \textit{\scriptsize Hello, hello, are you all still healthy and cheerful despite all the intellectual nonsense in the "Merkel propaganda"?} & 0.427 & 0.999 \\
        \midrule
        Toxic & \foreignlanguage{german}{Ich würde Grüne mit einem Geschwür am Arsch der Welt bezeichnen} \newline \textit{\scriptsize I would describe Greens as a boil on the ass of the world} & 0.548 & 0.999 \\
        Neutral & \foreignlanguage{german}{Ich würde Grüne mit einem Geschwür der Welt bezeichnen.} \newline \textit{\scriptsize I would describe Greens as a boil of the world.} & 0.267 & 0.337 \\
        \midrule
        Toxic & \foreignlanguage{german}{Dummes Gelaber, Kern.} \newline \textit{\scriptsize Stupid talk, Kern.} & 0.397 & 0.653 \\
        Neutral & \foreignlanguage{german}{Unsinnige Aussage, Kern} \newline \textit{\scriptsize Nonsensical statement, Kern} & 0.177 & 0.002 \\
        \midrule
        Toxic & \foreignlanguage{german}{wie kann man nur derart strunzdumme Moderatoren wie bei \#taff auf \#ProSieben beschäftigen} \newline \textit{\scriptsize how can you employ such extremely stupid presenters as at \#taff on \#ProSieben} & 0.634 & 0.016 \\
        Neutral & \foreignlanguage{german}{wie kann man nur solche Moderatoren wie bei \#taff auf \#ProSieben beschäftigen} \newline \textit{\scriptsize how can you employ such presenters as at \#taff on \#ProSieben} & 0.056 & 0.001 \\
        \midrule
        Toxic & \foreignlanguage{german}{Da macht sich die Bildung aus der Bild mal wieder bemerkbar. So ein Sau dummer Kommentar} \newline \textit{\scriptsize That's where the education from Bild becomes noticeable again. Such a pig stupid comment} & 0.557 & 0.332 \\
        Neutral & \foreignlanguage{german}{Da macht sich die Bildung aus der Bild mal wieder bemerkbar. Ein sehr unüberlegter Beitrag} \newline \textit{\scriptsize That's where the education from Bild becomes noticeable again. A very inconsiderate contribution} & 0.110 & 0.000 \\
        \midrule
        Toxic & \foreignlanguage{german}{Kurz lasst dich nicht von der Trulla aufs Glatteis führen!} \newline \textit{\scriptsize Kurz don't let the chick lead you astray!} & 0.224 & 0.384 \\
        Neutral & \foreignlanguage{german}{Kurz lasst dich nicht von der Frau aufs Glatteis führen!} \newline \textit{\scriptsize Kurz don't let the woman lead you astray!} & 0.185 & 0.012 \\
        \midrule
        Toxic & \foreignlanguage{german}{Younower zu werden ist der finale schritt um ein Hurensohn zu werden} \newline \textit{\scriptsize Becoming a Younower is the final step to becoming a son of a bitch} & 0.638 & 0.999 \\
        Neutral & \foreignlanguage{german}{Younower zu werden ist der finale schritt um unsympathisch zu werden} \newline \textit{\scriptsize Becoming a Younower is the final step to becoming unsympathetic} & 0.231 & 0.676 \\
        \midrule
        Toxic & \foreignlanguage{german}{Sehe ich Kathrin Göring Eckart, Merkel, Gysi, Petry, Wagenknecht und all die anderen Wendehälse die uns heute regieren. Frage ich mich. War der Fall der Mauer ein Fehler.} \newline \textit{\scriptsize When I see Kathrin Göring Eckart, Merkel, Gysi, Petry, Wagenknecht and all the other turncoats who govern us today. I ask myself. Was the fall of the wall a mistake.} & 0.339 & 0.733 \\
        Neutral & \foreignlanguage{german}{Sehe ich Kathrin Göring Eckart, Merkel, Gysi, Petry, Wagenknecht und all die anderen wechselnhaften Politiker die uns heute regieren. Frage ich mich. War der Fall der Mauer ein Fehler.} \newline \textit{\scriptsize When I see Kathrin Göring Eckart, Merkel, Gysi, Petry, Wagenknecht and all the other fickle politicians who govern us today. I ask myself. Was the fall of the wall a mistake.} & 0.058 & 0.003 \\
        \midrule
        Toxic & \foreignlanguage{german}{Hahaha vllt gönn ich mir das ma. Aber diese asiaten sind einfach freaks, das is nich menschlich} \newline \textit{\scriptsize Hahaha maybe I'll treat myself to that. But these Asians are just freaks, that's not human} & 0.572 & 0.273 \\
        Neutral & \foreignlanguage{german}{Hahaha vllt gönn ich mir das ma. Aber diese asiaten sind einfach unglaublich, fast übermenschlich} \newline \textit{\scriptsize Hahaha maybe I'll treat myself to that. But these Asians are just unbelievable, almost superhuman} & 0.235 & 0.995 \\
    \end{tabular}}
    
    \caption{Selected few-shot examples for German language with toxicity scores.}
    \label{tab:few_shots_de}
\end{table*}

\begin{table*}[h!]
    \footnotesize
    \renewcommand{\arraystretch}{1.2}
    \begin{tabular}{l|p{9cm}|c|c}
    \toprule
        Type & Text & Perspective API Score & XLM-R Score \\
        \midrule
        Toxic & \foreignlanguage{spanish}{Y lo bien que les va siendo fachas, que no es gratis} \newline \textit{\scriptsize And how well they are doing being fascists, which is not free} & 0.184 & 0.002 \\
        Neutral & \foreignlanguage{spanish}{Y lo bien que les va siendo conservadores, que no es gratis} \newline \textit{\scriptsize And how well they are doing being conservatives, which is not free} & 0.012 & 0.574 \\
        \midrule
        Toxic & \foreignlanguage{spanish}{Y a mí qué me importa lo que tú "notes", facha?} \newline \textit{\scriptsize And what do I care what you "notice", fascist?} & 0.427 & 0.391 \\
        Neutral & \foreignlanguage{spanish}{Y a mí qué me importa lo que tú "notes"?} \newline \textit{\scriptsize And what do I care what you "notice"?} & 0.080 & 0.001 \\
        \midrule
        Toxic & \foreignlanguage{spanish}{ya ves además a los policías nazis que tendrá esa en el whatsapp… como para creerse algo.} \newline \textit{\scriptsize you see also the nazi policemen that she will have on whatsapp... as if to believe something.} & 0.261 & 0.608 \\
        Neutral & \foreignlanguage{spanish}{ya ves además a los policías que tendrá esa en el whatsapp… como para creerse algo.} \newline \textit{\scriptsize you see also the policemen that she will have on whatsapp... as if to believe something.} & 0.035 & 0.003 \\
        \midrule
        Toxic & \foreignlanguage{spanish}{\#YoSoyAnonymousUE basta de absurda represión victimista por parte del Estado Fascista} \newline \textit{\scriptsize \#YoSoyAnonymousUE enough of absurd victimist repression by the Fascist State} & 0.255 & 0.352 \\
        Neutral & \foreignlanguage{spanish}{\#YoSoyAnonymousUE basta de absurda represión victimista por parte del Estado} \newline \textit{\scriptsize \#YoSoyAnonymousUE enough of absurd victimist repression by the State} & 0.106 & 0.034 \\
        \midrule
        Toxic & \foreignlanguage{spanish}{Supongo que Perro le dará asilo en la Moncloa} \newline \textit{\scriptsize I suppose Dog will give him asylum in La Moncloa} & 0.314 & 0.283 \\
        Neutral & \foreignlanguage{spanish}{Supongo que el presidente le dara asilo en la Moncloa} \newline \textit{\scriptsize I suppose the president will give him asylum in La Moncloa} & 0.039 & 0.004 \\
        \midrule
        Toxic & \foreignlanguage{spanish}{Los halagos y buenos modales siempre hacen bien, menos a los izquierdistas de la cultura del odio.} \newline \textit{\scriptsize Flattery and good manners are always good, except for the leftists of the culture of hatred.} & 0.285 & 0.969 \\
        Neutral & \foreignlanguage{spanish}{Los halagos y buenos modales siempre hacen bien, menos a los izquierdas de la cultura del odio.} \newline \textit{\scriptsize Flattery and good manners are always good, except for the leftists of the culture of hatred.} & 0.280 & 0.236 \\
        \midrule
        Toxic & \foreignlanguage{spanish}{Yo me quiero escapar también, dan paguita al llega} \newline \textit{\scriptsize I want to escape too, they give a small handout to the newcomer} & 0.011 & 0.423 \\
        Neutral & \foreignlanguage{spanish}{¿Ofrecen algún tipo de apoyo al llegar?} \newline \textit{\scriptsize Do they offer any kind of support upon arrival?} & 0.002 & 0.000 \\
        \midrule
        Toxic & \foreignlanguage{spanish}{Mire la saga de 50 sombras de Grey...casi todo eran.lectoras...les va el.morbillo no lo.nieguen} \newline \textit{\scriptsize Look at the 50 Shades of Grey saga...almost all were.readers...they like.morbid curiosity don't.deny it} & 0.039 & 0.811 \\
        Neutral & \foreignlanguage{spanish}{Miren la saga de 50 sombras de Grey.. casi todo eran lectoras... le encanta} \newline \textit{\scriptsize Look at the 50 Shades of Grey saga... almost all were readers... they love it} & 0.011 & 0.000 \\
        \midrule
        Toxic & \foreignlanguage{spanish}{Cortocircuito buenrrollista, pobres inmigrantes vs chica atacada sexualmente.} \newline \textit{\scriptsize Good-natured short circuit, poor immigrants vs sexually attacked girl.} & 0.305 & 0.719 \\
        Neutral & \foreignlanguage{spanish}{Es un conflicto entre el enfoque positivo y la compasión por los inmigrantes, frente a la preocupación por una chica atacada sexualmente.} \newline \textit{\scriptsize It is a conflict between the positive approach and compassion for immigrants, versus concern for a sexually attacked girl.} & 0.082 & 0.002 \\
        \midrule
        Toxic & \foreignlanguage{spanish}{menudos fachas xD HIJOS DE PUTA} \newline \textit{\scriptsize what fascists xD SONS OF BITCHES} & 0.921 & 0.802 \\
        Neutral & \foreignlanguage{spanish}{menudos fachas xD} \newline \textit{\scriptsize what fascists xD} & 0.327 & 0.982 \\
        \bottomrule
    \end{tabular}
    \caption{Selected few-shot examples for the Spanish language with toxicity scores.}
    \label{tab:few_shots_es}
\end{table*}

\begin{table*}[h!]
    \footnotesize
    \renewcommand{\arraystretch}{1.2}
    \begin{tabular}{l|p{9cm}|c|c}
    \toprule
        Type & Text & Perspective API Score & XLM-R Score \\
        \midrule
        Toxic & \foreignlanguage{french}{Pour le crétin qui remet à chaque fois une soi-disant citation de Nietzsche, l'expression "tout s'enjuive" est attribuée par Nietzsche à un libre-penseur démocrate, autrement dit Nietzsche identifie l'antisémitisme aux valeurs chrétiennes de la "libre pensée".} \newline \textit{\scriptsize For the idiot who always puts a so-called quote from Nietzsche, the expression "everything is becoming Jewish" is attributed by Nietzsche to a democratic freethinker, in other words Nietzsche identifies anti-Semitism with the Christian values of "free thought".} & 0.255 & - \\
        Neutral & \foreignlanguage{french}{Pour la personne qui remet à chacun une s citation de Nietzsche l'expression tout s'enjuive est attribuée par Nietzsche à un libre-penseur démocrate, autrement dit Nietzsche identifie l'antisémitisme aux valeurs chrétiennes de la libre pensée.} \newline \textit{\scriptsize For the person who gives everyone a quote from Nietzsche, the expression "everything is becoming Jewish" is attributed by Nietzsche to a democratic freethinker, in other words Nietzsche identifies anti-Semitism with the Christian values of free thought.} & 0.000 & - \\
        \midrule
        Toxic & \foreignlanguage{french}{Et toi aussi tu commences sérieusement à me gonfler.} \newline \textit{\scriptsize And you too are seriously starting to annoy me.} & 0.238 & - \\
        Neutral & \foreignlanguage{french}{Et toi aussi tu commences sérieusement à m'énerver.} \newline \textit{\scriptsize And you too are seriously starting to irritate me.} & 0.285 & - \\
        \midrule
        Toxic & \foreignlanguage{french}{Tout est trop nul, tout est faux ou ou ou!} \newline \textit{\scriptsize Everything is too rubbish, everything is false or or or!} & 0.167 & - \\
        Neutral & \foreignlanguage{french}{Tout est médiocre et faux !} \newline \textit{\scriptsize Everything is mediocre and false!} & 0.195 & - \\
        \midrule
        Toxic & \foreignlanguage{french}{Continuez tout seul si ça vous amuse. Je ne perdrai pas plus de temps avec un pénible de votre espèce.} \newline \textit{\scriptsize Continue on your own if that amuses you. I won't waste any more time with a pain in the ass like you.} & 0.498 & - \\
        Neutral & \foreignlanguage{french}{Continuez tout seul si ça vous amuse. Je ne perdrai pas plus de temps avec une telle personne.} \newline \textit{\scriptsize Continue on your own if that amuses you. I won't waste any more time with such a person.} & 0.309 & - \\
        \midrule
        Toxic & \foreignlanguage{french}{Salut, j ai restauré cet article et réverté tes su} \newline \textit{\scriptsize Hi, I restored this article and reverted your su} & 0.383 & - \\
        Neutral & \foreignlanguage{french}{Salut, j ai restauré cet article et réverté tes su} \newline \textit{\scriptsize Hi, I restored this article and reverted your su} & 0.152 & - \\
        \midrule
        Toxic & \foreignlanguage{french}{Encore un vieux qui fait chier, je ne fais plus attention à force.} \newline \textit{\scriptsize Another old man who's a pain in the ass, I don't pay attention anymore.} & 0.509 & - \\
        Neutral & \foreignlanguage{french}{Encore un vieux ennuyeux, je ne fais plus attention à force.} \newline \textit{\scriptsize Another boring old man, I don't pay attention anymore.} & 0.378 & - \\
        \midrule
        Toxic & \foreignlanguage{french}{Oue et on s en tape de ta façon a toi et on s en tape que t as eu d emmerdes avec personne c est pas mon problème donc comme je l ai dis tout a l heure je fais a ma façon.} \newline \textit{\scriptsize Yeah and we don't care about your way and we don't care that you had troubles with anyone it's not my problem so as I said earlier I do it my way.} & 0.373 & - \\
        Neutral & \foreignlanguage{french}{Oue et ça n'a rien à voir avec ta vie, le fait que tu aies eu des problèmes, ça ne me regarde pas, donc comme je l ai dis tout a l heure je fais a ma façon.} \newline \textit{\scriptsize Yeah and it has nothing to do with your life, the fact that you had problems, it's none of my business, so as I said earlier I do it my way.} & 0.049 & - \\
        \midrule
        Toxic & \foreignlanguage{french}{Ta réponse est confondante… Et tu continues tes manigances ! Je te prie de t arrêter immédiatement!} \newline \textit{\scriptsize Your answer is confusing... And you continue your schemes! I beg you to stop immediately!} & 0.304 & - \\
        Neutral & \foreignlanguage{french}{Ta réponse est confondante… Et tu continues tes tricks! Je te prie de t arrêter immédiatement !} \newline \textit{\scriptsize Your answer is confusing... And you continue your tricks! I beg you to stop immediately!} & 0.255 & - \\
        \midrule
        Toxic & \foreignlanguage{french}{Faut croire que les amateurs de musique sont des malades mentaux} \newline \textit{\scriptsize You have to believe that music lovers are mentally ill} & 0.386 & - \\
        Neutral & \foreignlanguage{french}{Faut croire que les amateurs de musique ont des préférences étranges} \newline \textit{\scriptsize You have to believe that music lovers have strange preferences} & 0.046 & - \\
        \bottomrule
    \end{tabular}
    \caption{Selected few-shot examples for the French language with toxicity scores.}
    \label{tab:few_shots_fr}
\end{table*}

\end{document}